\documentclass[runningheads]{llncs}
\usepackage{graphicx}
\usepackage{comment}
\usepackage{amsmath,amssymb} 
\usepackage{color}

\usepackage{microtype}
\usepackage[american]{babel}
\usepackage{bm}
\usepackage{multirow}
\usepackage{amsfonts}
\usepackage{mathrsfs}
\usepackage{epstopdf}
\usepackage{booktabs}
\usepackage{bbm}
\usepackage{dsfont}
\usepackage{makecell}
\usepackage{subfigure}
\usepackage{paralist}
\usepackage{ulem}

\newcommand{\CUB}{{\textit{CUB}}\xspace}
\newcommand{\Aircraft}{{\textit{Aircraft}}\xspace}
\newcommand{\NABirds}{{\textit{NABirds}}\xspace}
\newcommand{\VegFru}{{\textit{VegFru}}\xspace}
\newcommand{\Food}{{\textit{Food101}}\xspace}

\usepackage{array}
\newcommand{\PreserveBackslash}[1]{\let\temp=\\#1\let\\=\temp}
\newcolumntype{C}[1]{>{\PreserveBackslash\centering}p{#1}}
\newcolumntype{R}[1]{>{\PreserveBackslash\raggedleft}p{#1}}

\usepackage[linesnumbered,ruled,vlined]{algorithm2e}
\SetAlFnt{\scriptsize \sf}

\SetCommentSty{mycommfont}

\def\A{{\boldsymbol A}}
\def\a{{\boldsymbol a}}

\def\C{{\boldsymbol C}}
\def\c{{\boldsymbol c}}

\def\E{{\boldsymbol E}}
\def\F{{\boldsymbol F}}

\def\F{{\boldsymbol F}}
\def\f{{\boldsymbol f}}

\def\p{{\boldsymbol p}}

\def\X{{\boldsymbol X}}

\def\Q{{\boldsymbol Q}}

\def\S{{\boldsymbol S}}
\def\x{{\boldsymbol x}}
\def\y{{\boldsymbol y}}

\def\U{{\boldsymbol U}}
\def\u{{\boldsymbol u}}
\def\V{{\boldsymbol V}}
\def\v{{\boldsymbol v}}
\def\W{{\boldsymbol W}}

\def\X{{\boldsymbol X}}

\def\0{{\bf 0}}
\def\1{{\bf 1}}

\def\BM{{\mathcal B}}
\def\CM{{\mathcal C}}

\def\EM{{\mathcal E}}
\def\FM{{\mathcal F}}

\def\XM{{\mathcal X}}

\def\XM{{\mathcal X}}

\def\GM{{\mathcal G}}

\def\RB{{\mathbb R}}

\def\LM{{\mathcal L}}

\def\sq{\mathtt{sq}}	
\def\sp{\mathtt{sp}}	
\def\cp{\mathtt{cp}}	
	
\def\cat{\mathtt{cat}}	
\def\const{\mathtt{const}}	
\def\softmax{\mathtt{softmax}}	
\def\gap{\mathtt{GAP}}	
\def\sgn{\mathtt{sign}}
\def\tanh{\mathtt{tanh}}

\def\st{\mathtt{s.t.}}
\def\LFR{\mathtt{LFR}}	
\def\GFR{\mathtt{GFR}}	
\def\ie{\text{i.e.}}
\def\eg{\text{e.g.}}

\def\gap{\mathtt{GAP}}

\def\tr{\mathrm{tr}}

\begin{document}
\pagestyle{headings}
\mainmatter
\def\ECCVSubNumber{6502}  

\title{ExchNet: A Unified Hashing Network for Large-Scale Fine-Grained Image Retrieval} 


\titlerunning{ExchNet: A Unified Hashing Network for Large-Scale Fine-Grained Retrieval}

\author{Quan Cui$^\dag$ \inst{1,3} \and
Qing-Yuan Jiang$^\dag$ \inst{2} \and
Xiu-Shen Wei$^*$ \inst{3} \and
Wu-Jun Li\inst{2} \and
Osamu Yoshie\inst{1}}
\authorrunning{Q. Cui, Q.-Y. Jiang, X.-S. Wei, W.-J. Li, and O. Yoshie}
%
\institute{Graduate School of IPS, Waseda University, Japan \and
National Key Laboratory for Novel Software Technology, Department of Computer Science and Technology, Nanjing University, China \and
Megvii Research Nanjing, Megvii Technology, China\\
\email{cui-quan@toki.waseda.jp}, \email{qyjiang24@gmail.com}, \email{weixs.gm@gmail.com}, \email{liwujun@nju.edu.cn}, \email{yoshie@waseda.jp}}

\maketitle

\renewcommand{\thefootnote}{\dag}
\footnotetext[1]{Equal contribution.}
\renewcommand{\thefootnote}{*}
\footnotetext[2]{Corresponding author.}

\begin{abstract}
Retrieving content relevant images from a large-scale fine-grained dataset could suffer from intolerably slow query speed and highly redundant storage cost, due to high-dimensional real-valued embeddings which aim to distinguish subtle visual differences of fine-grained objects. In this paper, we study the novel fine-grained hashing topic to generate compact binary codes for fine-grained images, leveraging the search and storage efficiency of hash learning to alleviate the aforementioned problems. Specifically, we propose a unified end-to-end trainable network, termed as ExchNet. Based on attention mechanisms and proposed attention constraints, it can firstly obtain both local and global features to represent object parts and whole fine-grained objects, respectively. Furthermore, to ensure the discriminative ability and semantic meaning's consistency of these part-level features across images, we design a local feature alignment approach by performing a feature exchanging operation. Later, an alternative learning algorithm is employed to optimize the whole ExchNet and then generate the final binary hash codes. 
Validated by extensive experiments, our proposal consistently outperforms state-of-the-art generic hashing methods on five fine-grained datasets, which shows our effectiveness. Moreover, compared with other approximate nearest neighbor methods, ExchNet achieves the best speed-up and storage reduction, revealing its efficiency and practicality.
\end{abstract}

\section{Introduction}
Fine-Grained Image Retrieval (FGIR)~\cite{fgis_tmm,crl,towardscrl,scda,part_based_fgir,adversarial_fgir} is a practical but challenging computer vision task. It aims to retrieve images belonging to various sub-categories of a certain meta-category (\eg, birds, cars and aircrafts) and return images with the same sub-category as the query image. 
In real FGIR applications, previous methods could suffer from slow query speed and redundant storage costs due to both the explosive growth of massive fine-grained data and high-dimensional real-valued features.

In the literature, learning to hash~\cite{LSH:conf/compgeom/DatarIIM04,ITQ:conf/cvpr/GongL11,SH:conf/nips/WeissTF08,CNNH:conf/aaai/XiaPLLY14,DH:conf/cvpr/LiongLWMZ15,DPSH:conf/ijcai/LiWK16,DSH:conf/cvpr/Liu0SC16,DSDH:conf/nips/LiSHT17,DCH:conf/cvpr/CaoLL018,ADSH:conf/aaai/JiangL18,HashGAN:conf/cvpr/DizajiZSYDH18} has proven to be a promising solution for large-scale image retrieval because it can greatly reduce the storage costs and increase the query speeds. As a representative research area of approximate nearest neighbor~(ANN) search~\cite{LSH:conf/compgeom/DatarIIM04,PQ:journals/pami/JegouDS11,KDTree:journals/cacm/Bentley75}, hashing aims to embed data points as \mbox{similarity-preserving} binary codes. Recently, hashing has been successfully applied in a wide range of image retrieval tasks, e.g., face image retrieval~\cite{DDH:conf/ijcai/LinLT17}, person re-identification~\cite{PDH:journals/tip/ZhuKZFT17,CSBT:conf/cvpr/ChenWQLS17}, etc. We hereby explore the effectiveness of hashing for \textit{fine-grained} image retrieval.


\begin{figure*}[t]
\centering
\includegraphics[scale=0.34]{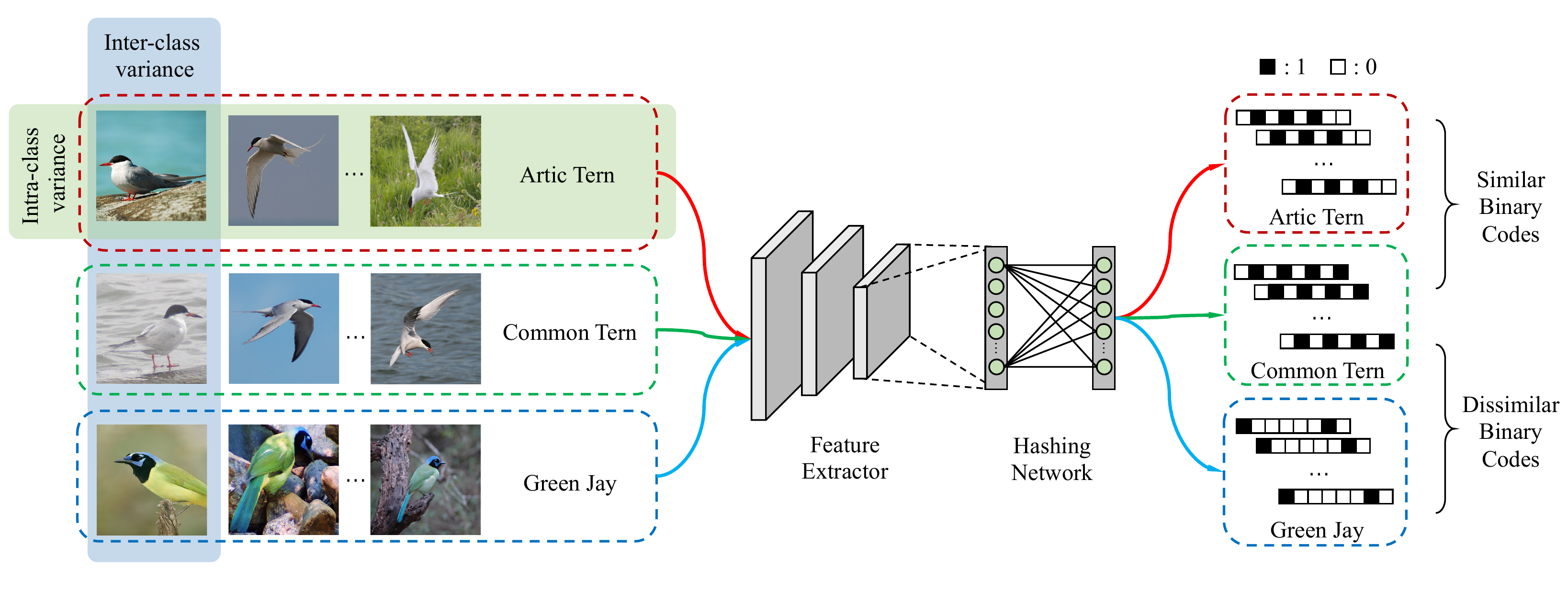}
\caption{Illustration of the fine-grained hashing task. Fine-grained images could share large intra-class variances but small inter-class variances. Fine-grained hashing aims to generate compact binary codes with tiny Hamming distances for images of the same sub-category, as well as distinct codes for images from different sub-categories.}
\label{fig:fg_hashing}
\end{figure*}

To the best of our knowledge, this is the first work to study the fine-grained hashing problem, which refers to the problem of designing hashing for fine-grained objects. As shown in Figure~\ref{fig:fg_hashing}, the task is desirable to generate compact binary codes for fine-grained images sharing both large intra-class variances and small inter-class variances. To deal with the challenging task, we propose a unified end-to-end trainable network ExchNet to first learn fine-grained tailored features and then generate the final binary hash codes.

\begin{figure*}[t]
\centering
\includegraphics[width=1.0\textwidth]{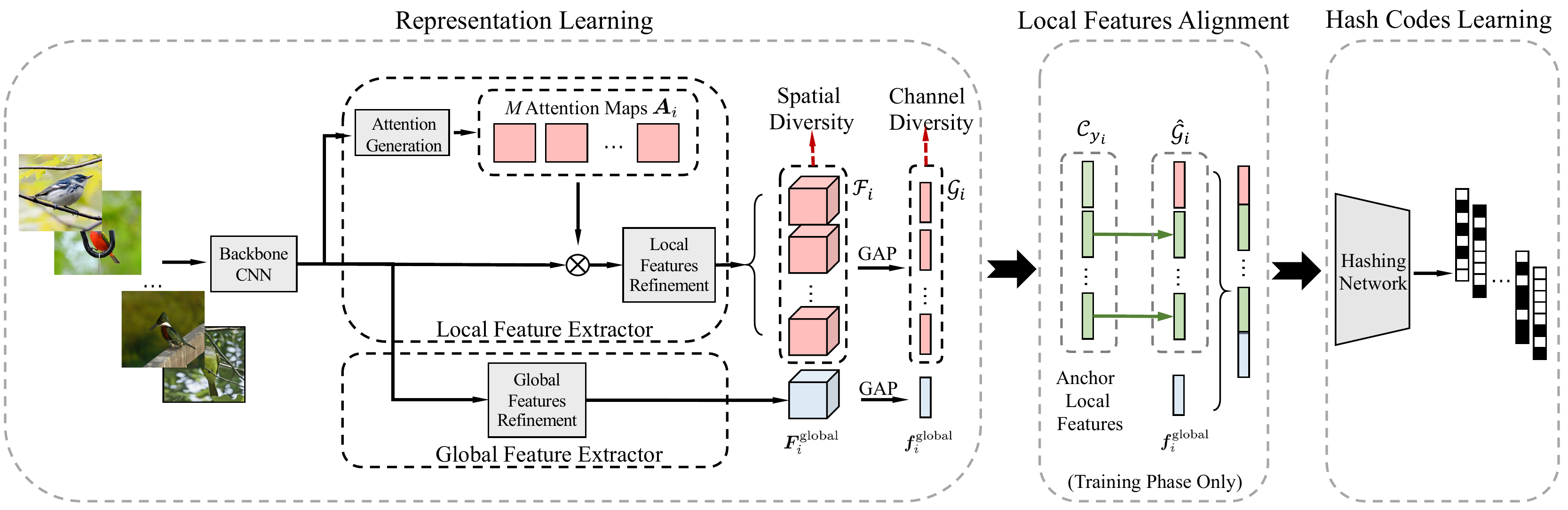}
\caption{Framework of our proposed ExchNet, which consists of three modules. 1) The representation learning module, as well as the attention mechanism with spatial and channel diversity learning constraints, is designed to obtain both local and global features of fine-grained objects. 2) The local feature alignment module is used to align obtained local features w.r.t. object parts across different fine-grained images. 3) The hash codes learning module is performed to generate the compact binary codes.}
\label{fig:framework}
\end{figure*}

In concretely, our ExchNet consists of three main modules, including representation learning, local feature alignment and hash code learning, as shown in Figure~\ref{fig:framework}. In the representation learning module, beyond obtaining the holistic image representation (i.e., global features), we also employ the attention mechanism to capture the part-level features (i.e., local features) for representing fine-grained objects' parts. Localizing parts and embedding part-level cues are crucial for fine-grained tasks, since these discriminative but subtle parts (e.g., bird heads or tails) play a major role to distinguish different sub-categories. Moreover, we also develop two kinds of attention constraints, i.e., spatial and channel constraints, to collaboratively work together for further improving the discriminative ability of these local features. In the following, to ensure that these part-level features can correspond to their own corresponding parts across different fine-grained images, we design an anchor based feature alignment approach to align these local features. Specifically, in the local feature alignment module, we treat the anchored local features as the ``prototype'' w.r.t. its sub-category by averaging all the local features of that part across images. Once local features are well aligned for their own parts, even if we exchange one specific part's local feature of an input image with the same part's local feature of the prototype, the image meanings derived from the image representations and also the final hash codes should be both extremely similar. Inspired by this motivation, we perform a feature exchanging operation upon the anchored local features and other learned local features, which is illustrated in Figure~\ref{fig:keyidea}. After that, for effectively training the network with our feature alignment fashion, we utilize an alternating algorithm to solve the hashing learning problem and update anchor features simultaneously.



%
%

To quantitatively prove both effectiveness and efficiency of our ExchNet, we conduct comprehensive experiments on five fine-grained benchmark datasets, including the large-scale ones, i.e., \NABirds~\cite{NABirds:conf/cvpr/HornBFHBIPB15}, \VegFru~\cite{VegFru:conf/iccv/HouFW17} and \Food~\cite{Food101:conf/eccv/BossardGG14}. Particularly, compared with competing approximate nearest neighbor methods, our ExchNet achieves up to hundreds times speedup for large-scale fine-grained image retrieval without significant accuracy drops. Meanwhile, compared with state-of-the-art generic hashing methods, ExchNet could consistently outperform these methods by a large margin on all the fine-grained datasets. Additionally, ablation studies and visualization results justify the effectiveness of our tailored model designs like local feature alignment and proposed attention approach.


The contributions of this paper are summarized as follows:
\begin{itemize}
	\item We study the novel fine-grained hashing topic to leverage the search and storage efficiency of hash codes for solving the challenging large-scale fine-grained image retrieval problem.
	\item We propose a unified end-to-end trainable network, i.e., ExchNet, to first learn fine-grained tailored features and then generate the final binary hash codes. Particularly, the proposed attention constraints, local feature alignment and anchor-based learning fashion contribute well to obtain discriminative fine-grained representations.
	\item We conduct extensive experiments on five fine-grained datasets to validate both effectiveness and efficiency of our proposed ExchNet. Especially for the results on large-scale datasets, ExchNet exhibits its outperforming retrieval performance on either speedup, memory usages and retrieval accuracy.
\end{itemize}

\section{Related Work}

\subsubsection{Fine-Grained Image Retrieval}
%
Fine-Grained Image Retrieval (FGIR) is an active research topic emerged in recent years, where the database and query images could share small inter-class variance but large intra-class variance. 
In previous works~\cite{fgis_tmm}, handcrafted features were initially utilized to tackle the FGIR problem. Powered by deep learning techniques, more and more deep learning based FGIR methods~\cite{fgis_tmm,crl,maskcnn,towardscrl,scda,part_based_fgir,adversarial_fgir,piecewise} were proposed. These deep methods can be roughly divided into two parts, \ie, supervised and unsupervised methods.
In supervised methods, FGIR is defined as a metric learning problem. Zheng et al.~\cite{crl} designed a novel ranking loss and a weakly-supervised attractive feature extraction strategy to facilitate the retrieval performance. Zheng et al.~\cite{towardscrl} improved their former work~\cite{crl} with a normalize-scale layer and de-correlated ranking loss.  
As to unsupervised methods, Selective Convolutional Descriptor Aggregation (SCDA)~\cite{scda} was proposed to localize the main object in fine-grained images firstly, and then discard the noisy background and keep useful deep descriptors for fine-grained image retrieval. 


\subsubsection{Deep Hashing}
Hashing methods can be divided into two categories, i.e., data-independent methods~\cite{LSH:conf/compgeom/DatarIIM04} and data-dependent methods~\cite{ITQ:conf/cvpr/GongL11,DPSH:conf/ijcai/LiWK16}, based on whether training points are used to learn hash functions. Generally speaking, data-dependent methods, also named as Learning to Hash~(L2H) methods, can achieve better retrieval performance with the help of the learning on training data. With the rise of deep learning, some L2H methods integrate deep feature learning into hash frameworks and achieve promising performance. As previous work, many deep hashing methods~\cite{CNNH:conf/aaai/XiaPLLY14,DH:conf/cvpr/LiongLWMZ15,DPSH:conf/ijcai/LiWK16,DSH:conf/cvpr/Liu0SC16,DSDH:conf/nips/LiSHT17,DCH:conf/cvpr/CaoLL018,ADSH:conf/aaai/JiangL18,HashGAN:conf/cvpr/DizajiZSYDH18,eccv:hashing1,eccv:hashing2,eccv:hashing3,eccv:hashing4,eccv:hashing5} for large-scale image retrieval have been proposed. Compared with deep unsupervised hashing methods~\cite{DH:conf/cvpr/LiongLWMZ15,HashGAN:conf/cvpr/DizajiZSYDH18,ADSH:conf/aaai/JiangL18}, deep supervised hashing methods~\cite{CNNH:conf/aaai/XiaPLLY14,DPSH:conf/ijcai/LiWK16,DSDH:conf/nips/LiSHT17,ADSH:conf/aaai/JiangL18} can achieve superior retrieval accuracy as they can fully explore the semantic information. Specifically, the previous work~\cite{CNNH:conf/aaai/XiaPLLY14} was essentially a two-stage method which tried to learn binary codes in the first stage and employed feature learning guided by the learned binary codes in the second stage. Then, there appeared numerous one-stage deep supervised hashing methods, including Deep Pairwise Supervised Hashing~(DPSH)~\cite{DPSH:conf/ijcai/LiWK16}, Deep Supervised Hashing~(DSH)~\cite{DSH:conf/cvpr/Liu0SC16}, and Deep Cauchy Hashing~(DCH)~\cite{DCH:conf/cvpr/CaoLL018}, which aimed to integrate feature learning and hash code learning into an end-to-end framework. Hashing Network~(HashNet)~\cite{HashGAN:conf/cvpr/DizajiZSYDH18} utilized $\tanh(\beta x)$ to approximate $\sgn(x)$ by increasing $\beta$. Asymmetric Deep Supervised Hashing~(ADSH)~\cite{ADSH:conf/aaai/JiangL18} tried to use asymmetric hashing improve the training efficiency and retrieval performance.

\section{Methodology}

The framework of our ExchNet is presented in Figure~\ref{fig:framework}, which contains three key modules, i.e., the representation learning module, local feature alignment module, and hash code learning module. 


\begin{figure*}[t]
\centering
\includegraphics[scale=0.35]{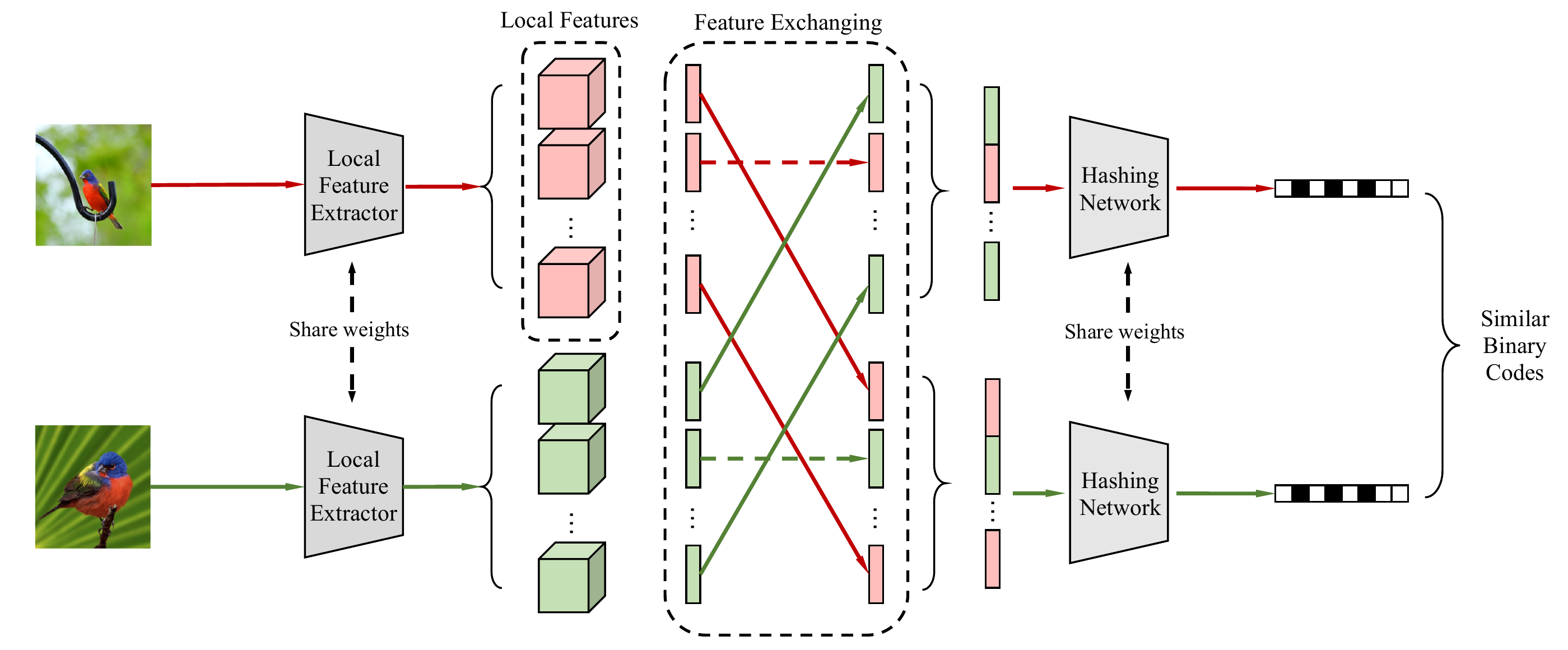}
\caption{Key idea of our local feature alignment approach: Given an image pair of a fine-grained category, exchanging their local features of the same object parts should not change their corresponding hash codes, \ie, these hash codes should be the same as those generated without local feature exchanging and their Hamming distance should be still close also.}
\label{fig:keyidea}
\end{figure*}


\subsection{Representation Learning}

The learning of discriminative and meaningful local features is mutually correlated with fine-grained tasks~\cite{bcnn,isqrt,navigate,macnn,racnn}, since these local features can greatly benefit the distinguishing of sub-categories with subtle visual differences deriving from the discriminative fine-grained parts ({e.g.}, bird heads or tails) .
In consequence, as shown in Figure~\ref{fig:framework}, beyond the global feature extractor, we also introduce a local feature extractor in the representation learning module. Specifically, by considering model efficiency,  we hereby propose to learn local features with the attention mechanism, rather than other fine-grained techniques with tremendous computation cost, \eg, second-order representations~\cite{bcnn,isqrt} or complicated network architectures~\cite{navigate,macnn,racnn}.
 
Given an input image $\x_i$, a backbone CNN is utilized to extract a holistic deep feature $\E_i\in\RB^{H\times W\times C}$, which serves as the appetizer for both the local feature extractor and the global feature extractor.

It is worth mentioning that the attention is engaged in the middle of the feature extractor. Since, in the shallow layers of deep neural networks, low-level context information~(\eg, colors and edges, etc.) are well preserved, which is crucial for distinguish subtle visual differences of fine-grained objects. Then, by feeding $\E_i$ into the attention generation module, $M$ pieces of attention maps $\A_i\in\RB^{M\times H\times W}$ are generated and we use $\A^j_i\in\RB^{H\times W}$ to denote the attentive region of the $j$-th~($j\in\{1,\ldots,M\}$) part cues for $\x_i$. After that, the obtained part-level attention map $\A^j_i$ is element-wisely multiplied on $\E_i$ to select the attentive local feature corresponding to the $j$-th part, which is formulated as:
\begin{align}
\hat\E^j_i=\E_i\otimes\A^j_i,
\end{align}
where $\hat\E^j_i\in\RB^{H\times W\times C}$ represents the $j$-th attentive local feature of $\x_i$, and ``$\otimes$'' denotes the Hadamard product on each channel. For simplification, we use $\hat\EM_i=\{\hat\E^1_i,\dots,\hat\E^M_i\}$ to denote a set of local features and, subsequently, $\hat\EM_i$ is fed into the later Local Features Refinement~(LFR) network composed of a stack of convolution layers to embed these attentive local features into higher-level semantic meanings:
\begin{align}
\FM_i = f_{\LFR}(\hat\EM_i),
\end{align}
where the output of the network is denoted as $\FM_i=\{\F_i^1,\dots,\F_i^M\}$, which represents the final local feature maps w.r.t. high-level semantics. We denote $\f_i^j\in\RB^{C'}$ as the local feature vector after applying global average pooling~(GAP) on $\F_i^j\in\RB^{H'\times W'\times C'}$ as:
\begin{align}
\f_i^j = f_\gap(\F_i^j)\,.
\end{align}

On the other side, as to the global feature extractor, for $\x_i$, we directly adopt a Global Features Refinement~(GFR) network composed of conventional convolutional operations to embed $\E_i$, which is presented by:
\begin{align}
\F^{\rm global}_i = f_{\GFR}(\E_i)\,.
\end{align}
We use $\F^{\rm global}_i\in\RB^{H'\times W'\times C'}$ and $\f^{\rm global}_i\in\RB^{C'}$ to denote the learned global feature and the corresponding holistic feature vector after GAP, respectively.

Furthermore, to facilitate the learning of localizing local feature cues (i.e., capturing fine-grained parts), we impose the spatial diversity and channel diversity constraints over the local features in $\FM_i$.

Specifically, it is a natural choice to increase the diversity of local features by differentiating the distributions of attention maps~\cite{macnn}. However, it might cause a problem that the holistic feature can not be activated in some spatial positions, while the attention map has large activation values on them due to over-applied constraints upon the learned attention maps. Instead, in our method, we design and apply constraints on the local features. In concretely, for the local feature $\F^j_i$, we obtain its ``aggregation map'' $\hat\A_i^j\in\RB^{H'\times W'}$ by adding all $C'$ feature maps through the channel dimension and apply the softmax function on it for converting it into a valid distribution, then flat it into a vector $\hat\a_i^j$. Based on the Hellinger distance, we propose a spatial diversity induced loss as:
\begin{align}
\LM_{\mathtt{sp}}(\x_i)=1-\frac{1}{\sqrt{2}\binom{M}{2}}\sum_{l,k=1}^M\left\Vert\sqrt{\hat\a_i^l}-\sqrt{\hat\a_i^k}\right\Vert_2	,\label{eq:sp-loss}
\end{align}
where $\binom{M}{2}$ is used to denote the combinatorial number of ways to pick $2$ unordered outcomes from $M$ possibilities. The spatial diversity constraint drives the aggregation maps to be activated in spatial positions as diverse as possible.
As to the channel diversity constraint, we first convert the local feature vector $\f_i^j$ into a valid distribution, which can be formulated by
\begin{align}
&\p_i^j=\softmax(\f_i^j),\;\forall j\in\{1,\dots,M\}.
\end{align}
Subsequently, we propose a constraint loss over $\{\p_i^j\}_{j=1}^M$ as:
\begin{align}
\LM_{\mathtt{cp}}(\x_i)=\left[t-\frac{1}{\sqrt{2}\binom{M}{2}}\sum_{l,k=1}^M\left\Vert\sqrt{\p_i^l}-\sqrt{\p_i^k}\right\Vert_2\right]_+	,\label{eq:cp-loss}
\end{align}
where $t\in[0,1]$ is a hyper-parameter to adjust the diversity and $[\cdot]_+$ denotes $max(\cdot, 0)$. Equipping with the channel diversity constraint could benefit the network to depress redundancies in features through channel dimensions. Overall, our spatial diversity and channel diversity constraints can work in a collaborative way to obtain discriminative local features.

\begin{figure*}[t]
\centering
\includegraphics[scale=0.35]{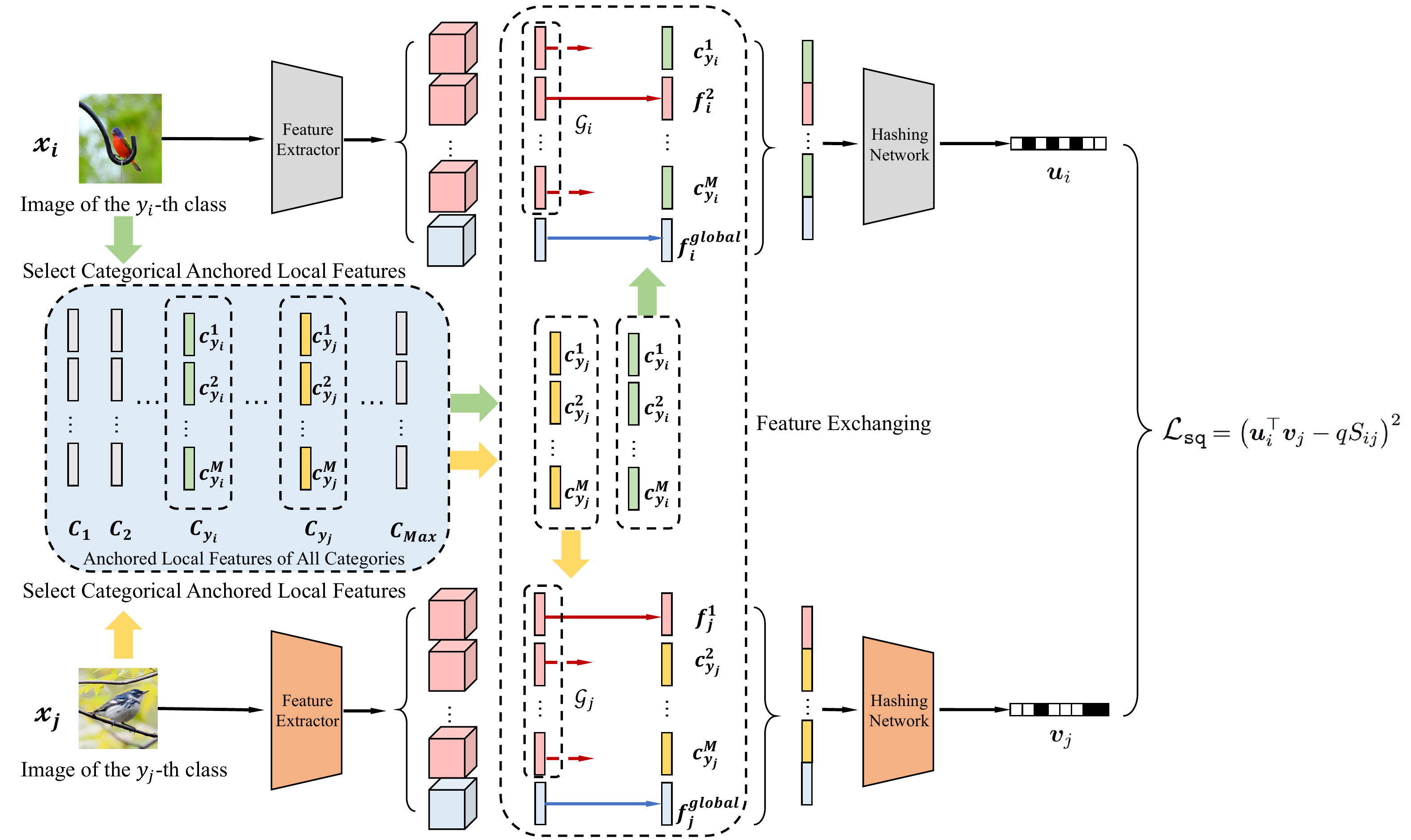}
\caption{Our feature exchanging and hash codes learning in the training phase. According to the class indices~(\ie, $y_{i}$ and $y_{j}$), we first select categorical anchor features $\CM_{y_i}$ and $\CM_{y_j}$ for samples $\x_i$ and $\x_j$, respectively. Then, for each input image, the feature exchanging operation is conducted between its learned and anchored local features. After that, hash codes are generated with exchanged features and the learning is driven by preserving pairwise similarities of hash codes $\u_i$ and $\v_j$.}
\label{fig:exchange}
\end{figure*}

\subsection{Learning to Align by Local Feature Exchanging}\label{sec:learning_to_align}

Upon the representation learning module, the alignment on local features is necessary for confirming that they represent and more importantly correspond to common fine-grained parts across images, which are essential to fine-grained tasks.
Hence, we propose an anchor-based local features alignment approach assisted with our feature exchanging operation.

Intuitively, local features from the same object part (e.g., bird heads of a bird species) should be embedded with almost the same semantic meaning. As illustrated by Figure~\ref{fig:keyidea}, our key idea is that, if local features were well aligned, exchanging the features of identical parts for two input images belonging to the same sub-category should not change the generated hash codes. Inspired by that, we propose a local feature alignment strategy by leveraging the feature exchanging operation, which happens between learned local features and anchored local features. 
As a foundation for feature exchanging, a set of dynamic anchored local features $\CM_{y_i}=\{\c^1_{y_i},\dots,\c^M_{y_i}\}$ for class $y_i$ should be maintained, in which the $j$-th anchored local feature $\c^j_{y_i}$ is obtained by averaging all $j$-th part's local features of training samples from class $y_i$. At the end of each training epoch, anchored local features will be recalculated and updated.
Subsequently, as shown in Figure~\ref{fig:exchange}, for a sample $\x_i$ whose category is $y_i$, we exchange a half of the learned local features in $\GM_i=\{\f_i^1,\dots,\f_i^M\}$ with its corresponding anchored local features in $\CM_{y_i}=\{\c^1_{y_i},\dots,\c^M_{y_i}\}$. The exchanging process can be formulated as:
\begin{align}
\forall j\in\{1,\dots,M\},\hat\f_i^j\triangleq
\left\{
\begin{aligned}
\f_i^j,\quad\text{if }\xi_j\ge 0.5,\\
\c^j_{y_i},\quad\text{otherwise,}
\end{aligned} 
\right.
\end{align}
where $\xi_j\sim\BM(0.5)$ is a random variable following the Bernoulli distribution for the $j$-th part. The local features after exchanging are denoted as $\hat\GM_i=\{\hat\f_i^1,\dots,\hat\f_i^M\}$ and fed into the hashing learning module for generating binary codes and computing similarity preservation losses.

\subsection{Hash Code Learning}

After obtaining both global features and local features, we concatenate them together and feed them into the hashing learning module. Specifically, the hashing network contains a fully connected layer and a $\sgn(\cdot)$ activation function layer. In our method, we choose an asymmetric hashing for ExchNet for its flexibility~\cite{powerofasym}. Concretely, we utilize two hash functions, defined as $g(\cdot)$ and $h(\cdot)$, to learn two different binary codes for the same training sample. The learning procedure is as follows:
\begin{align}
\u_i=g([\hat\GM_{i};\f^{\rm global}_i]_{\cat})=\sgn(\W^{(g)}[\hat\GM_{i};\f^{\rm global}_i]_{\cat}),\\
\v_i=h([\hat\GM_{i};\f^{\rm global}_i]_{\cat})=\sgn(\W^{(h)}[\hat\GM_{i};\f^{\rm global}_i]_{\cat}),
\end{align}
where $[;]_{\rm{cat}}$ denotes the concatenation operator, and $\u_i,\v_i\in\{-1,+1\}^{q}$ denote the two different binary codes of the $i$-th sample. $q$ represents the code length. $\W^{(g)}$ and $\W^{(h)}$ present the parameters of hash functions $g(\cdot)$ and $h(\cdot)$\footnote{\scriptsize We omit the bias term for simplicity.}, respectively. We denote $\U=\{\u_i\}_{i=1}^{n}$ and $\V=\{\v_i\}_{i=1}^{n}$ as learned binary codes. Inspired by~\cite{ADSH:conf/aaai/JiangL18}, we only keep binary codes $\v_i$ and set hash function $h(\cdot)$ implicitly. Hence, we can perform feature learning and binary codes learning simultaneously.

To preserve the pairwise similarity, we adopt the squared loss and define the following objective function:
\begin{align}
\LM_{\sq}(\u_i,\v_j,\bm{\CM})=\big(\u_i^\top\v_j-qS_{ij}\big)^2,
\end{align}
where $\u_i=g([\hat\GM_{i};\f^{\rm global}_i]_\cat)$, $S_{ij}$ is the pairwise similarity label and $\bm{\CM}=\{\CM_i\}_{i=1}^M$. We use $\Theta$ to denote the parameters of deep neural network and hash layer. The aforementioned process is generally illustrated by Figure~\ref{fig:exchange}.

Due to the zero-gradient problem caused by the $\sgn(\cdot)$ function, $L_\sq(\cdot,\cdot,\cdot)$ becomes intractable to optimize. In this paper, we relax $g(\cdot) = \sgn(\cdot)$ into $\tilde g (\cdot)=\tanh(\cdot)$ to alleviate this problem. Then, we can derive the following loss function:
\begin{align}
\tilde \LM_\sq(\tilde\u_i,\v_j,\bm{\CM})=\big(\tilde\u_i^\top\v_j-qS_{ij}\big)^2,\label{eq:sq-loss}
\end{align}
where $\tilde\u_i=\tilde g([\hat\GM_i;\f^{\rm global}_i]_\cat)$ and $\U$ becomes $\tilde\U=\{\tilde\u_i\}_{i=1}^n$.

Then, given a set of image samples $\XM=\{\x_1,\dots,\x_n\}$ and their pairwise labels $\S=\{S_{ij}\}_{i,j=1}^n$, we can get the following objective function by combining Equation~(\ref{eq:sp-loss}),~(\ref{eq:cp-loss}) and~(\ref{eq:sq-loss}):
\begin{align}\label{pro:obj}
\min_{\V,\Theta,\bm{\CM}}&\;\LM(\XM)=\sum_{i,j=1}^n\tilde \LM_{\sq}(\tilde\u_i,\v_j;S_{ij})+\lambda\sum_{i=1}^n\LM_{\sp}(\x_i)+\gamma\sum_{i=1}^n\LM_{\cp}(\x_i)\\
\st\ &\forall i\in\{1,\dots,n\},\hat\u_i=\hat g([\hat\GM_{i};\f^{\rm global}_i]_\cat),\v_j\in\{-1,+1\}^q,\nonumber
\end{align}
where $S_{ij}$ represents the similarity between the $i$-th and $j$-th samples, $q$ denotes the code length, $\lambda$ and $\gamma$ are hyper-parameters.

\subsection{Learning Algorithm}

To solve the optimization problem in Equation~(\ref{pro:obj}), we design an alternating algorithm to learn $\V$, $\Theta$, and $\bm{\CM}$. Specifically, we learn one parameter with the others fixed. 


\subsubsection{Learn $\Theta$ with $\V$ and $\CM$ fixed}
When $\V$, $\CM$ fixed, we use back-propagation~(BP) to update the parameters $\Theta$. In particular, for input sample $\x_i$, we first calculate the following gradient:
\begin{align}
\nabla_\Theta\LM(\X)=	\sum_{i,j=1}^n\nabla_\Theta \LM_\sq(\hat\u_i,\v_j)+\lambda\sum_{i=1}^n\nabla_\Theta \LM_\sp(\x_i)+\gamma\sum_{i=1}^n\nabla_\Theta \LM_\cp(\x_i).\label{sol:graident}
\end{align}
Then, we use the back-propagation algorithm to update $\Theta$.

\subsubsection{Learn $\V$ with $\Theta$ and $\CM$ fixed}
When $\Theta$, $\CM$ are fixed, we rewrite $\LM(\V)$ as follows:
\begin{align}
\LM(\V)
&=\sum_{i,j=1}^n\big(\hat\u_i^\top\v_j-qS_{ij}\big)^2=\Vert\widetilde\U\V^\top-q\S\Vert^2_F\\
&=\Vert\widetilde\U\V^\top\Vert^2_F-2q\tr(\S^\top\widetilde\U\V^\top)+\const.
\end{align}

Because $\V$ is defined over $\{-1,+1\}^{n\times q}$, we learn $\V$ column by columns as that in ADSH~\cite{ADSH:conf/aaai/JiangL18}. Specifically, we can get the closed-form solution for the $k$-th column $\V_{*k}$ as follows:
\begin{align}
\V_{*k}=\sgn(\V_{/k}\widetilde\U_{/k}^\top\widetilde\U_{*k}-2q\Q_{*k}),\label{sol:Vk}
\end{align}
where $\Q=\S^\top\widetilde\U$.

\subsubsection{Learn $\bm{\CM}$ with $\V$ and $\Theta$ fixed}
When $\Theta$, $\V$ fixed, we use the following equation to update each $\CM_i\in\bm{\CM}$:
\begin{align}
\forall k,\c^k_i=\frac{1}{n_i}\sum_{i=1}^{n_{i}}\f^k_i,\label{sol:Ci}
\end{align}
where $n_i$ denotes the number of samples in class $y_i$.

\subsection{Out-of-Sample Extension}
When we finish the training phase, we can generate the binary code for the sample $\x_i$ by $\u_i=\sgn(\W^{(g)}[\GM_{i};\f^{\rm global}_i]_\cat)$.

\section{Experiments}

\subsection{Datasets}
For comparisons, we select two widely used fine-grained datasets, \ie, \CUB~\cite{CUB:journals/CalTech/WBSWPB2011} and \Aircraft~\cite{AirCraft:journals/corr/MajiRKBV13}, as well as three popular large-scale fine-grained datasets, \ie, \NABirds~\cite{NABirds:conf/cvpr/HornBFHBIPB15}, \VegFru~\cite{VegFru:conf/iccv/HouFW17}, and \Food~\cite{Food101:conf/eccv/BossardGG14}, to conduct experiments.

Specifically, \CUB is a bird classification benchmark dataset containing $11,788$ images from 200 bird species. It is officially split into $5,994$ for training and $5,794$ for test. 
\Aircraft contains $10,000$ images from $100$ kinds of aircraft model variants with $6667$ for training and $3333$ for test. 
Moreover, for large-scale datasets, \NABirds has $555$ common species of birds in North America with $23,929$ training images and $24,633$ test images.
\VegFru is a large-scale fine-grained dataset covering vegetables and fruits from $292$ categories with $29,200$ for training and $116,931$ for test.
\Food contains $101$ kinds of foods with $101,000$ images. For each class, $250$ test images are manually reviewed for correctness while $750$ training images still contain some amount of noises.

\subsection{Baselines and Implementation Details}
\subsubsection{Baselines}
For comparisons with other ANN algorithms, we select two tree-based ANN methods, i.e., BallTree~\cite{BallTree:journals/corr/DolatshahHM15} and KDTree~\cite{KDTree:journals/cacm/Bentley75}, and one production quantization based ANN method, i.e., Product Quantization~(PQ)~\cite{PQ:journals/pami/JegouDS11}. The linear scan means that we directly perform exhaustive search based on the learned real-valued features. 
For comparisons with other hashing baselines, we choose eight state-of-the-art generic hashing methods. They are LSH~\cite{LSH:conf/compgeom/DatarIIM04}, SH~\cite{SH:conf/nips/WeissTF08}, ITQ~\cite{ITQ:conf/cvpr/GongL11}, SDH~\cite{SDH:conf/cvpr/ShenSLS15}, DPSH~\cite{DPSH:conf/ijcai/LiWK16}, DSH~\cite{DSH:conf/cvpr/Liu0SC16}, HashNet~\cite{Hashnet:conf/iccv/CaoLWY17}, and ADSH~\cite{ADSH:conf/aaai/JiangL18}. Among these methods, DPSH, DSH, HashNet and ADSH are based on deep learning and others are not.

\subsubsection{Implementation Details}
For comparisons with other ANN algorithms, we carry out experiments on \Food in which the database is the largest. We first utilize the triplet loss~\cite{tripletloss} to learn $512$-D and $1024$-D feature embeddings for its frequent usages in fine-grained retrieval tasks. Then, the performance of linear scan is tested on the learned features. More experimental settings about BallTree~\cite{BallTree:journals/corr/DolatshahHM15}, KDTree~\cite{KDTree:journals/cacm/Bentley75} and PQ~\cite{PQ:journals/pami/JegouDS11} can be found in the supplementary materials. 
For our ExchNet, the retrieval procedure is divided into coarse ranking to select top~$N$ as candidates and re-ranking to return top~$K$~($K<N$) from top~$N$ candidates. We adopt the real-valued features learned with the triplet loss directly. 
As presented in Table~\ref{tab:retrieval}, we report results including precision at top~$K$~(P@K), wall clock time~(WC time), speed up ratio, and memory cost.

Our backbone employs the first three stages of ResNet50 and the attention generation module is the fourth stage of ResNet50 without downsample convolutions. The LFR and GFR of ExchNet are independent networks, sharing the same architecture with the fourth stage of ResNet50. The optimizer is standard mini-batch stochastic gradient descent with weight decay $1\times 10^{-4}$. The mini-batch size $M$ is set to 64 and the iteration times $T_{max}$ is 100. Learning rate is set to 0.001, which is divided by 10 at the 60-th and 80-th iteration, respectively. The hyper-parameter $t$ is set to $0.4$. The number of training epochs is 20. For efficient training, we randomly sample a subset of the training set in each epoch. Specifically, for \CUB, \Aircraft, \Food, we sample 2,000 samples per epoch, while 4,000 samples are randomly selected for other datasets. To provide reliable local features for our local feature alignment strategy, in the first 50 iterations, since both local and global features are not well learned, the part-level feature exchanging operation is disabled for avoiding aligning meaningless local features. 

\subsection{Comparisons with other ANN Methods} \label{sec:speedup}

To prove the practicality and effectiveness of our proposed method, comparisons with other ANN methods are presented in this section. All experiments are conducted based on hash codes of $32$bits generated by our model. 

In Table~\ref{tab:retrieval}, we present the retrieval performance on the \Food dataset. Specifically, we present the P@10, WC time, speedup, and memory cost for all methods. 
We can observe that, compared with the linear search, our method can achieve up to $233\times$ and $395\times$ acceleration on features of $512$-D and $1024$-D, respectively. The memory cost of our method is also much less than tree-based methods. 
The best speed-up and the lowest storage usage prove the practicality of our proposed method. 
Meanwhile, our method can achieve state-of-the-art retrieval accuracies, which demonstrates that our ExchNet is the most effective one compared with other ANN methods. 
Above results illustrate our ExchNet deserves to be the optimal choice for fine-grained image retrieval.

\begin{table}[t]
\scriptsize
\centering
\caption{Retrieval performance comparisons on the \Food dataset.}
\setlength{\tabcolsep}{1pt}
\label{tab:retrieval}
\begin{tabular}{l|cccc|cccc}
\hline
\multirow{2}{*}{Method} &  \multicolumn{4}{c|}{512-dim}&  \multicolumn{4}{c}{1024-dim}\\\cline{2-9}
            &P@10($\uparrow$)           & WCtime($\downarrow$)         & Speedup($\uparrow$)              & Memory($\downarrow$) &P@10($\uparrow$)           & WCtime($\downarrow$)         & Speedup($\uparrow$)              & Memory($\downarrow$) \\ \hline
Linear & 80.05\%        &9,481.03       & $1\times$            &  207.2MB    & 80.28\%        &22,377.96      & $1\times$            &  414.1MB    \\ \hline
BallTree   & 77.22\%        &236.23         & $40.13\times$        &  28.1MB     & 77.74\%        &213.88         & $104.62\times$     &  28.1MB     \\
KDTree     & 77.42\%        &70.16          & $135.13\times$       &  28.8MB     & 77.73\%        &73.57          & $304.14\times$     &  28.7MB     \\
PQ          & 77.12\%        &43.49          & $217.99\times$       &  524.5KB    & 77.18\%        &72.47          & $308.74\times$     &  1.0MB      \\\hline
\textbf{Ours}         & \textbf{77.69\%}        &\textbf{40.54}          & $\bm{233.85}\times$       &  \textbf{404.0KB}    & \textbf{78.06\%}        &\textbf{56.57}          & $\bm{395.53}\times$     &  \textbf{404.0KB}    \\
\hline
\end{tabular}
\end{table}

\begin{table*}[t]
\scriptsize
\centering
\caption{Comparisons of retrieval accuracy~(MAP) on all the fine-grained datasets.}
\label{tab:acc-hash}
\begin{tabular}{C{1.3cm}|R{0.9cm}|R{0.9cm}|R{0.9cm}|R{0.9cm}|R{0.9cm}|R{0.9cm}|R{0.9cm}|R{1.1cm}|R{0.9cm}|R{0.9cm}}
\hline
Method    & \makecell[c]{\#Bits}  & \makecell[c]{LSH}& \makecell[c]{SH}&  \makecell[c]{ITQ}&  \makecell[c]{SDH}&  \makecell[c]{DPSH}&  \makecell[c]{DSH}& \makecell[c]{HashNet}& \makecell[c]{ADSH}& \makecell[c]{\textbf{Ours}}\\
\hline
\multirow{4}*{\CUB}
          & 12bits  & 2.26\%  & 5.55\%  & 6.80\%  & 10.52\%  & 8.68\%  & 4.48\% 	& 12.03\%  & 20.03\%  &{\bf 25.14\%}\\
          & 24bits  & 3.59\%  & 6.72\%  & 9.42\%  & 16.95\%  & 12.51\%  & 7.97\%  & 17.77\%  & 50.33\%  &{\bf 58.98\%}\\
          & 32bits  & 5.01\%  & 7.63\%  & 11.19\%  & 20.43\%  & 12.74\%  & 7.72\% 	& 19.93\%  & 61.68\%  &{\bf 67.74\%}\\
          & 48bits  & 6.16\%  & 8.32\%  & 12.45\%  & 22.23\%  & 15.58\%  & 11.81\%  & 22.13\%  & 65.43\%  &{\bf 71.05\%}\\
\hline
\multirow{4}*{\Aircraft}
          & 12bits  & 1.69\%  & 3.28\% 	& 4.38\%  & 4.89\% 	& 8.74\%  & 8.14\% 	& 14.91\%  & 15.54\% 	&{\bf 33.27\%}\\
          & 24bits  & 2.19\%  &	3.85\% 	& 5.28\%  &	6.36\% 	& 10.87\%  &	10.66\% 	& 17.75\%  &	23.09\% 	&{\bf 45.83\%}\\
          & 32bits  & 2.38\%  &	4.04\% 	& 5.82\%  &	6.90\% 	& 13.54\%  &	12.21\% 	& 19.42\%  &	30.37\% 	&{\bf 51.83\%}\\
          & 48bits  & 2.82\%  &	4.28\%	& 6.05\%  &	7.65\%  & 13.94\%  &	14.45\%  & 20.32\%  &	50.65\%  &{\bf 59.05\%}\\
\hline
\multirow{4}*{\NABirds}
          & 12bits  & 0.90\%  & 2.12\%  & 2.53\%  & 3.10\% 	& 2.17\%  & 1.56\% 	& 2.34\%  & 2.53\% 	&{\bf 5.22\%}\\
          & 24bits  & 1.68\%  & 3.14\% 	& 4.22\%  &	6.72\% 	& 4.08\%  &	2.33\% 	& 3.29\%  &	8.23\% 	&{\bf 15.69\%}\\
          & 32bits  & 2.43\%  & 3.71\% 	& 5.38\%  &	8.86\% 	& 3.61\%  &	2.44\% 	& 4.52\%  &	14.71\% 	&{\bf 21.94\%}\\
          & 48bits  & 3.09\%  &	4.05\% 	& 6.10\%  &	10.38\% 	& 3.20\%  &	3.42\% 	& 4.97\%  &	25.34\% 	&{\bf 34.81\%}\\
\hline
\multirow{4}*{\VegFru}
          & 12bits  & 1.28\%  & 2.36\% 	& 3.05\%  & 5.92\% 	& 6.33\%  & 4.60\% 	& 3.70\%  & 8.24\% 	&{\bf 23.55\%}\\
          & 24bits  & 2.21\%  &	4.04\% 	& 5.51\%  &	11.55\% 	& 9.05\%  &	8.91\% 	& 6.24\%  &	24.90\% 	&{\bf 35.93\%}\\
          & 32bits  & 3.39\%  &	5.65\% 	& 7.48\%  &	14.55\% 	& 10.28\%  &	11.23\% 	& 7.83\%  &	36.53\% 	&{\bf 48.27\%}\\
          & 48bits  & 4.51\%  &	6.56\% 	& 8.74\%  &	16.45\% 	& 9.11\%  &	17.12\% 	& 10.29\%  &	55.15\% 	&{\bf 69.30\%}\\
\hline
\multirow{4}*{\Food}
          & 12bits  & 1.57\%  & 4.51\% 	& 6.46\%  & 10.21\%  & 11.82\%  &	6.51\%  & 24.42\%  &	35.64\% 	&{\bf 45.63\%}\\
          & 24bits  & 2.48\%  &	5.79\%	& 8.20\%  &	11.44\% 	& 13.05\%  & 8.97\% 	& 34.48\%  &	40.93\% 	&{\bf 55.48\%}\\
          & 32bits  & 2.64\%  &	5.91\% 	& 9.70\%  &	13.36\% 	& 16.41\%  &	13.10\% 	& 35.90\%  &	42.89\% 	&{\bf 56.39\%}\\
          & 48bits  & 3.07\%  &	6.63\%  & 10.07\%  &	15.55\%  & 20.06\%  &	17.18\%  & 39.65\%  &	48.81\%  &{\bf 64.19\%}\\
\hline
\end{tabular}
\end{table*}

\subsection{Comparisons with State-of-the-art Hashing Methods}
In Table~\ref{tab:acc-hash}, we present the mean average precision~(MAP) results for comparisons with state-of-the-art hashing methods on all datasets. From Table~\ref{tab:acc-hash}, we can observe that our method can achieve the best retrieval performance in all cases. 
On fine-grained datasets~(\CUB and \Aircraft) of relatively small size, almost all the generic hashing methods (except for ADSH) can not achieve a satisfactory performance, \ie, a relatively low MAP. Also, our ExchNet outperforms the most powerful baseline ADSH considerably. 
It can verify that given limited training data, our proposed method could still perform well. 
As to large-scale fine-grained datasets, the improvements become more significant. 
Particularly, comparing with the most powerful baselines, we achieve $12\%$ and $14\%$ MAP improvements on the 32 bits and 48 bits evaluation experiments of the large-scale \VegFru dataset. Meanwhile, we achieve $14\%$ and $16\%$ MAP improvements on the 32 bits and 48 bits experiments of the \Food dataset. 
It shows that, with sufficient training data, we can get better retrieval results with our ExchNet on large-scale fine-grained datasets.

\subsection{Ablation Studies}
\subsubsection{Effectiveness of the Exchanging-based Feature Alignment}
\begin{figure*}[t]
\centering
\begin{tabular}{c@{ }@{ }c@{ }@{ }c@{ }@{ }c}
\begin{minipage}{0.32\linewidth}\centering
    \includegraphics[width=1\textwidth]{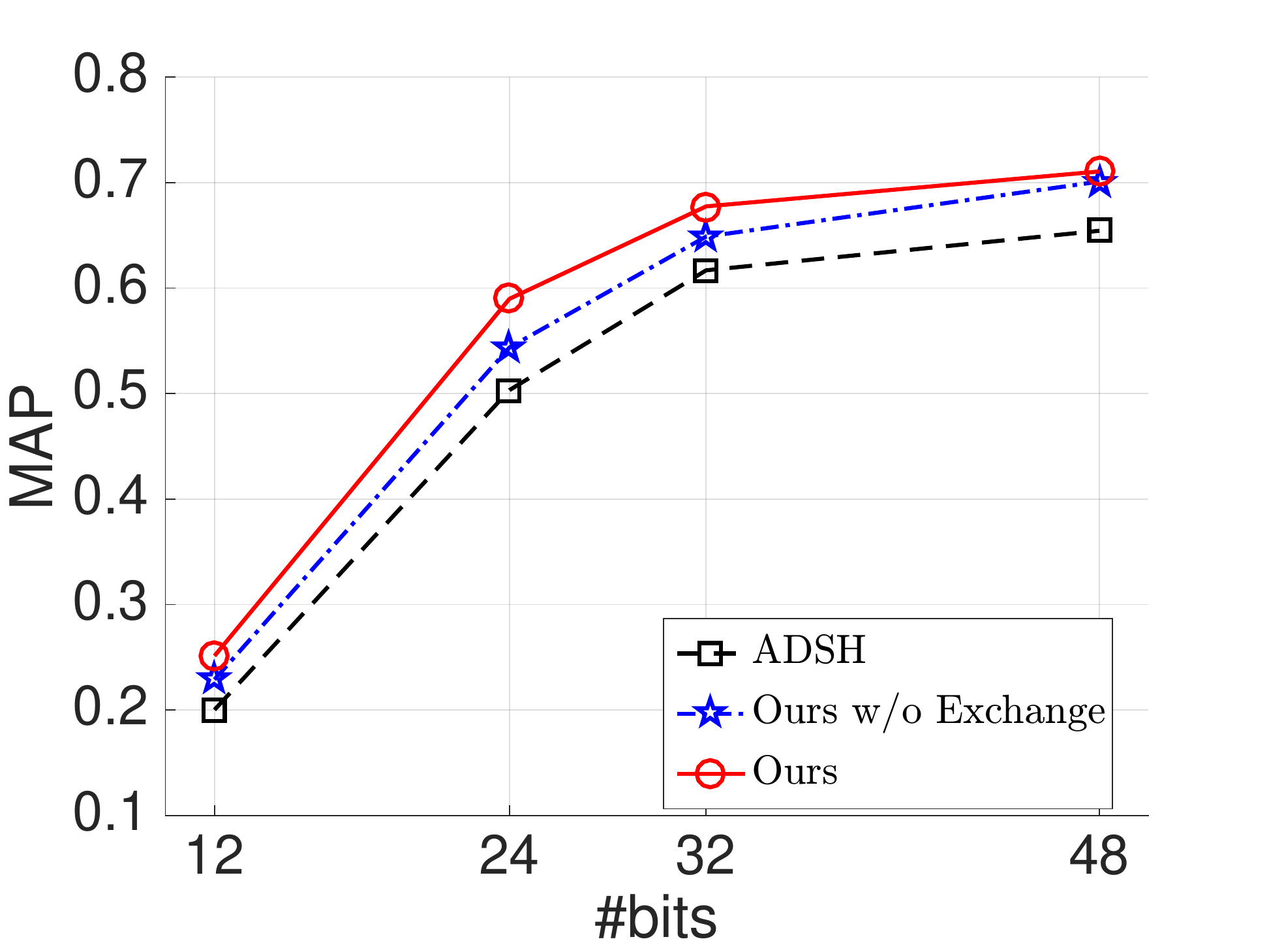}\\
    {\scriptsize(a) \CUB}
\end{minipage} &
\begin{minipage}{0.32\linewidth}\centering
    \includegraphics[width=1\textwidth]{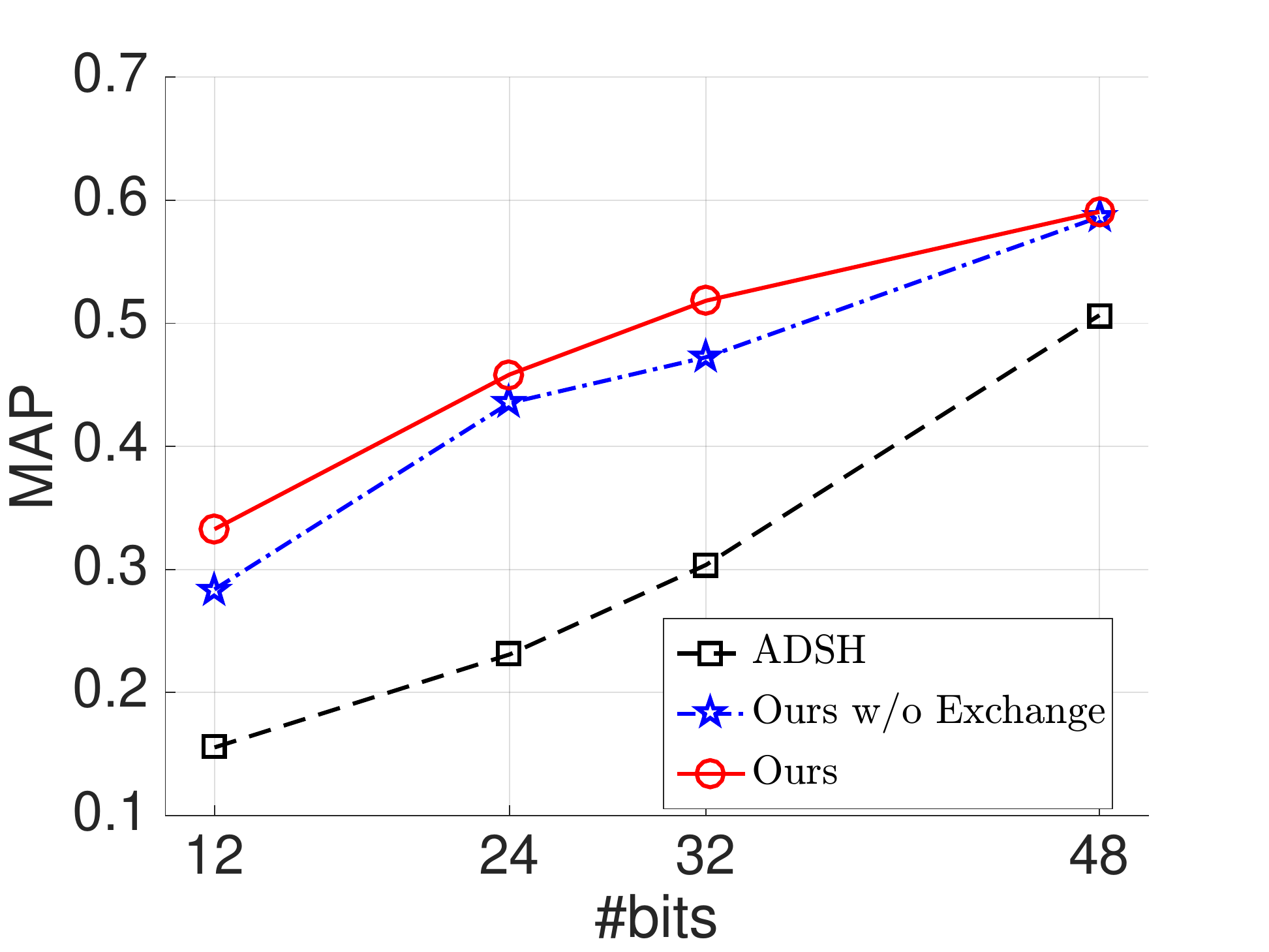}\\
    {\scriptsize(b) \Aircraft}
\end{minipage}&
\begin{minipage}{0.32\linewidth}\centering
    \includegraphics[width=1\textwidth]{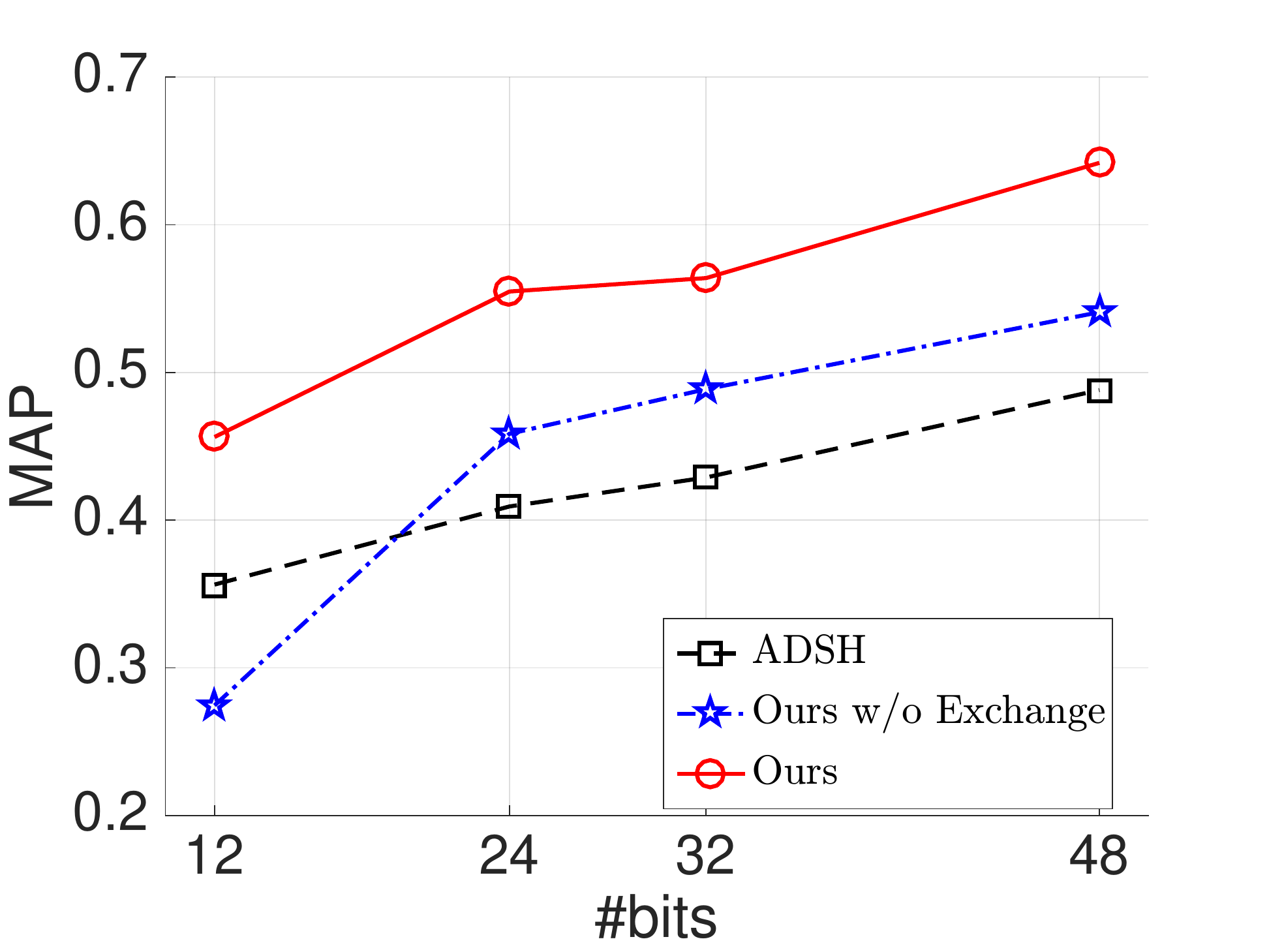}\\
    {\scriptsize(c) \Food}
\end{minipage}
\end{tabular}
\caption{Effectiveness of our feature exchanging operation.}
\label{fig:ablation}
\end{figure*}

We verify the effectiveness of the local feature alignment approach~(cf. Section~\ref{sec:learning_to_align}) in this section. The retrieval accuracy are present in Figure~\ref{fig:ablation}, where ``Ours w/o Exchange'' means that we do not perform the feature exchanging operation (\ie, the local feature alignment) during training. Note that ``Ours w/o Exchange'' is degenerated to the ADSH~\cite{ADSH:conf/aaai/JiangL18} learned with our proposed representation learning architecture instead of ResNet50. Hence, we also present the results of ADSH.

It can be observed that our method can achieve the best accuracy thanks to the feature exchanging operation. Specifically, on \CUB and \Aircraft datasets, our proposed method with the exchanging operation performs considerably better than that without exchanging. 
The performance improvement on the large-scale fine-grained datasets~(\eg, \Food) becomes more significant. 
Above results illustrate that our proposed local features alignment strategy is effective, especially on large-scale datasets. Moreover, even if bits of hash codes are limited, our feature alignment strategy could still benefit fine-grained retrieval greatly. 

\begin{figure*}[t]
\centering
\begin{tabular}{c@{ }@{ }c@{ }@{ }c@{ }@{ }c}
\begin{minipage}{0.32\linewidth}\centering
    \includegraphics[width=1\textwidth]{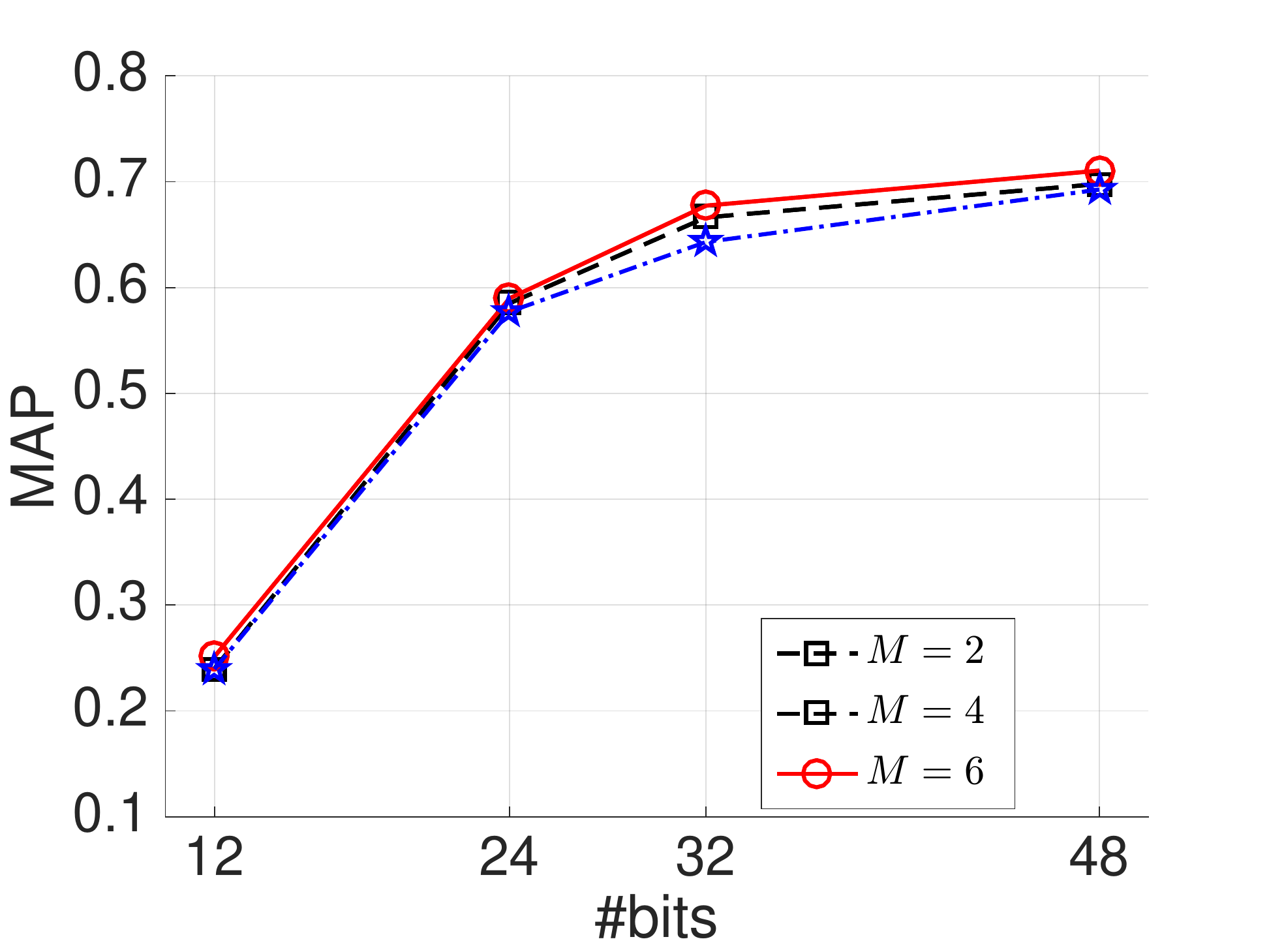}\\
    {\scriptsize(a) \CUB}
\end{minipage} &
\begin{minipage}{0.32\linewidth}\centering
    \includegraphics[width=1\textwidth]{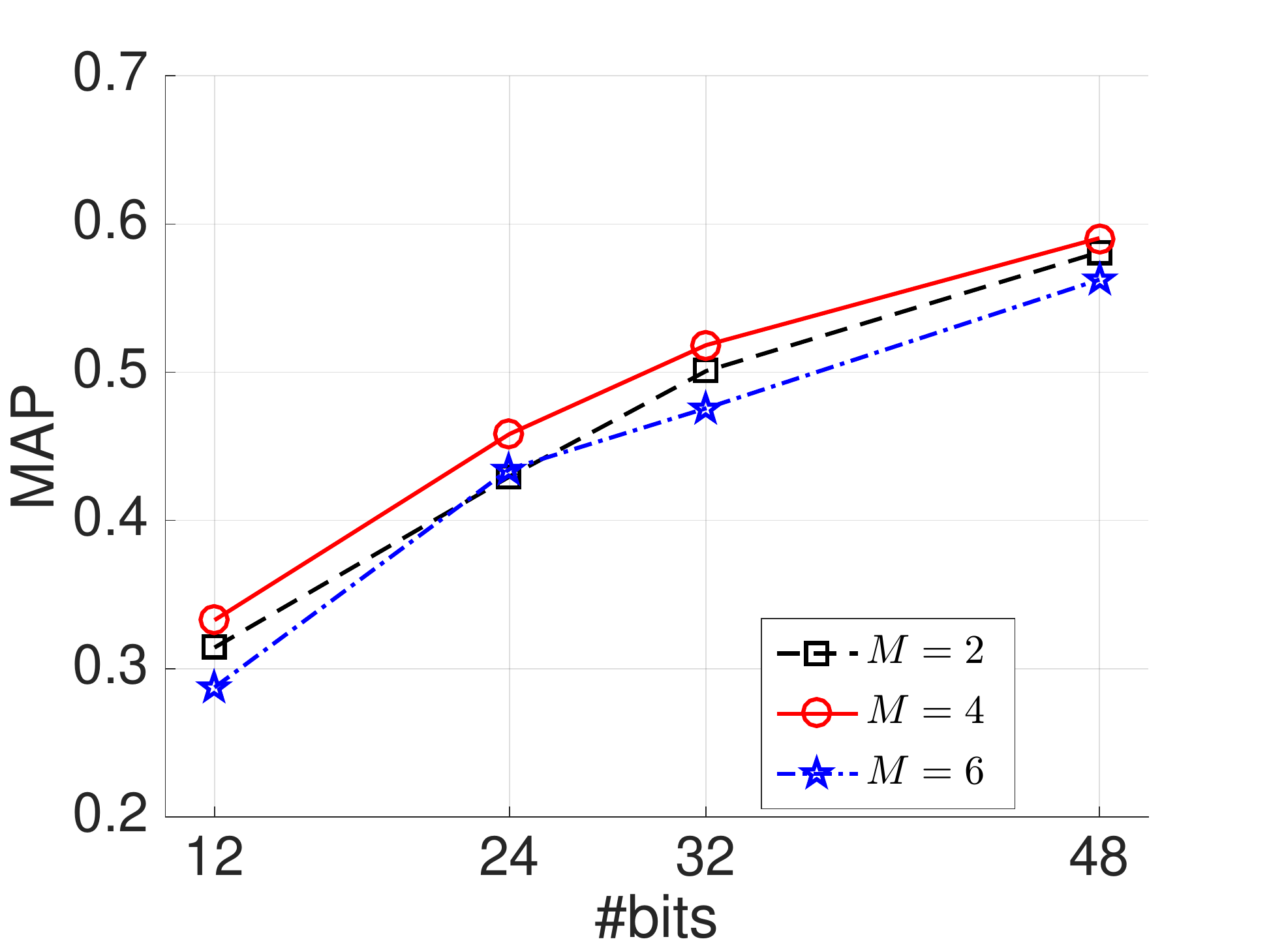}\\
    {\scriptsize(b) \Aircraft}
\end{minipage}&
\begin{minipage}{0.32\linewidth}\centering
    \includegraphics[width=1\textwidth]{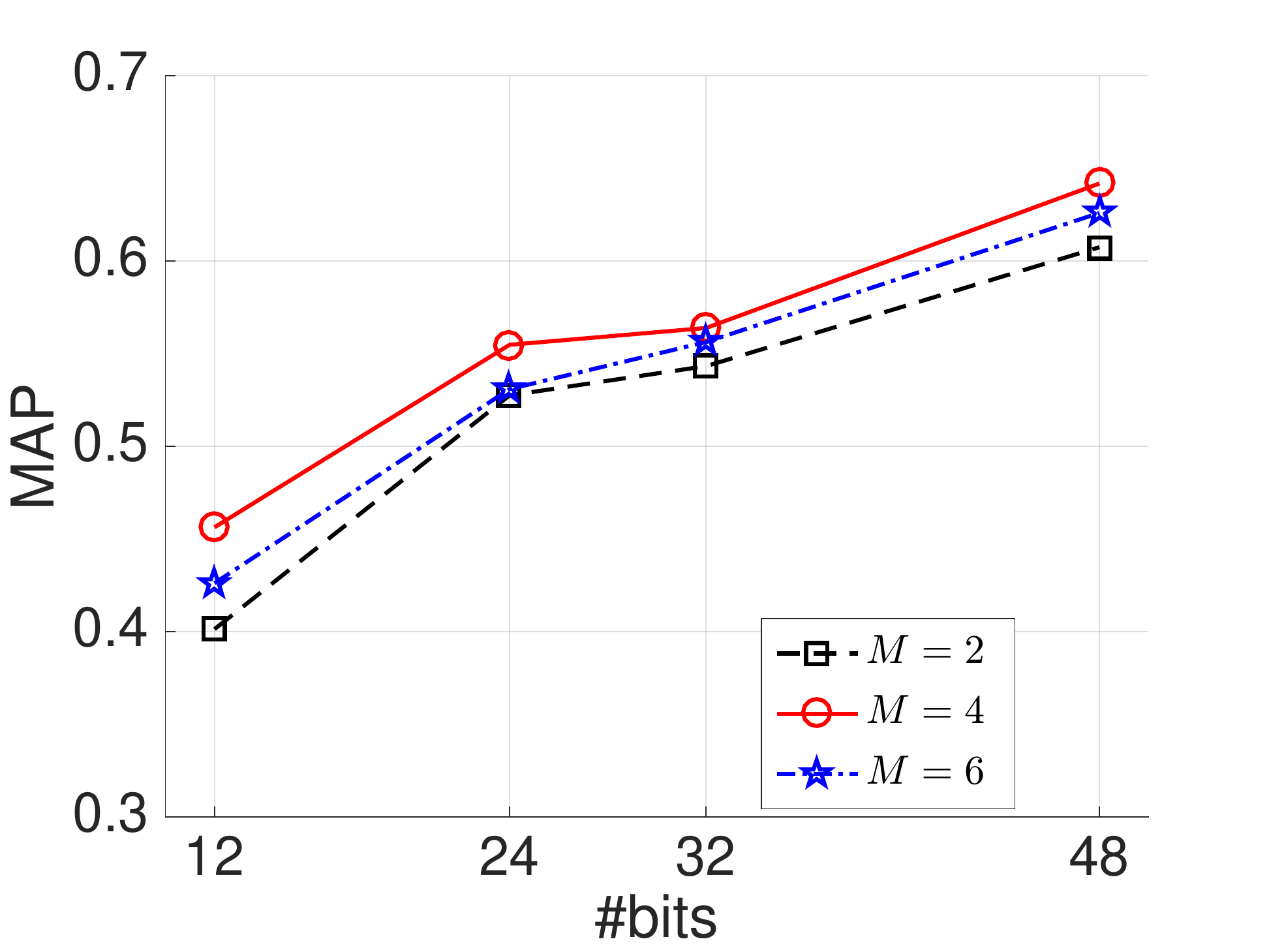}\\
    {\scriptsize(c) \Food}
\end{minipage}
\end{tabular}
\caption{Influence of hyper parameter $M$ which denotes the number of local features.}
\label{fig:M}
\end{figure*}

\subsubsection{Sensitivity to Hyper Parameter $M$}
In our ExchNet, we use $M$ to denote the number of local features, which is also the number of attention maps. In this section, we present the influence of the hyper-parameter $M$ by ablation studies.

As presented in Figure~\ref{fig:M}, we vary $M$ as $2$, $4$ and $6$. From that figure, it is observed that satisfactory retrieval accuracies are achieved regardless of different $M$ values, and the best fine-grained retrieval accuracy is obtained when $M=4$. As analyzed, redundant local features~(\ie, overmuch object parts when $M$ is large) might cause redundancies in local feature representations, while the lack of local features~(\ie, scant object parts when $M$ is small) may result in that fine-grained images are under-represented for distinguishing subtle visual differences. Those might be the reasons why $M$ is too small or large will cause slightly accuracy drops. Moreover, comparable retrieval results of different $M$ values show that our ExchNet is not sensitive to $M$.

\section{Conclusions}

In this paper, we studied the practical but challenging fine-grained hashing task, which aims to solve large-scale FGIR problems by leveraging the search and storage efficiency of compact hash codes. Specifically, we proposed a unified network ExchNet to obtain representative fine-grained local and global features by performing our attention approach equipped with the tailored attention constraints. Then, ExchNet utilized its local feature alignment to align these local features to their corresponding object parts across images. Later, an alternating learning algorithm was employed to return the final fine-grained binary codes. Compared with ANN methods and competing generic hash methods, experiments validated both effectiveness and efficiency of our ExchNet. In the future, we would like to explore a more challenging unsupervised fine-grained hashing topic. \\

\noindent\textbf{Acknowledgements} Q. Cui's contribution was made when he was an intern at Megvii Research Nanjing. This research was supported by the National Key Research and Development Program of China under Grant 2017YFA0700800 and ``111'' Program B13022.


\clearpage
%
%
\bibliographystyle{splncs04}
\bibliography{egbib}

\begin{thebibliography}{10}
\providecommand{\url}[1]{\texttt{#1}}
\providecommand{\urlprefix}{URL }
\providecommand{\doi}[1]{https://doi.org/#1}

\bibitem{KDTree:journals/cacm/Bentley75}
Bentley, J.L.: Multidimensional binary search trees used for associative
  searching. {ACM} Commun.  \textbf{18}(9),  509--517 (1975)

\bibitem{eccv:hashing2}
Cakir, F., He, K., Sclaroff, S.: Hashing with binary matrix pursuit. In: ECCV.
  pp. 332--348 (2018)

\bibitem{DCH:conf/cvpr/CaoLL018}
Cao, Y., Long, M., Liu, B., Wang, J.: Deep cauchy hashing for hamming space
  retrieval. In: CVPR. pp. 1229--1237 (2018)

\bibitem{Hashnet:conf/iccv/CaoLWY17}
Cao, Z., Long, M., Wang, J., Yu, P.S.: Hashnet: Deep learning to hash by
  continuation. In: ICCV. pp. 5609--5618 (2017)

\bibitem{CSBT:conf/cvpr/ChenWQLS17}
Chen, J., Wang, Y., Qin, J., Liu, L., Shao, L.: Fast person re-identification
  via cross-camera semantic binary transformation. In: CVPR. pp. 5330--5339
  (2017)

\bibitem{LSH:conf/compgeom/DatarIIM04}
Datar, M., Immorlica, N., Indyk, P., Mirrokni, V.S.: Locality-sensitive hashing
  scheme based on p-stable distributions. In: SoCG. pp. 253--262 (2004)

\bibitem{HashGAN:conf/cvpr/DizajiZSYDH18}
Dizaji, K.G., Zheng, F., Sadoughi, N., Yang, Y., Deng, C., Huang, H.:
  Unsupervised deep generative adversarial hashing network. In: CVPR. pp.
  3664--3673 (2018)

\bibitem{BallTree:journals/corr/DolatshahHM15}
Dolatshah, M., Hadian, A., Minaei{-}Bidgoli, B.: Ball*-tree: Efficient spatial
  indexing for constrained nearest-neighbor search in metric spaces. CoRR
  \textbf{abs/1511.00628} (2015)

\bibitem{racnn}
Fu, J., Zheng, H., Mei, T.: Look closer to see better: Recurrent attention
  convolutional neural network for fine-grained image recognition. In: CVPR.
  pp. 4438--4446 (2017)

\bibitem{ITQ:conf/cvpr/GongL11}
Gong, Y., Lazebnik, S.: Iterative quantization: {A} procrustean approach to
  learning binary codes. In: {CVPR}. pp. 817--824 (2011)

\bibitem{NABirds:conf/cvpr/HornBFHBIPB15}
Horn, G.V., Branson, S., Farrell, R., Haber, S., Barry, J., Ipeirotis, P.,
  Perona, P., Belongie, S.J.: Building a bird recognition app and large scale
  dataset with citizen scientists: The fine print in fine-grained dataset
  collection. In: CVPR. pp. 595--604 (2015)

\bibitem{VegFru:conf/iccv/HouFW17}
Hou, S., Feng, Y., Wang, Z.: Vegfru: {A} domain-specific dataset for
  fine-grained visual categorization. In: ICCV. pp. 541--549 (2017)

\bibitem{PQ:journals/pami/JegouDS11}
J{\'{e}}gou, H., Douze, M., Schmid, C.: Product quantization for nearest
  neighbor search. IEEE TPAMI  \textbf{33}(1),  117--128 (2011)

\bibitem{ADSH:conf/aaai/JiangL18}
Jiang, Q., Li, W.: Asymmetric deep supervised hashing. In: AAAI. pp. 3342--3349
  (2018)

\bibitem{isqrt}
Li, P., Xie, J., Wang, Q., Gao, Z.: Towards faster training of global
  covariance pooling networks by iterative matrix square root normalization.
  In: CVPR. pp. 947--955 (2018)

\bibitem{DSDH:conf/nips/LiSHT17}
Li, Q., Sun, Z., He, R., Tan, T.: Deep supervised discrete hashing. In:
  NeurIPS. pp. 2482--2491 (2017)

\bibitem{DPSH:conf/ijcai/LiWK16}
Li, W., Wang, S., Kang, W.: Feature learning based deep supervised hashing with
  pairwise labels. In: IJCAI. pp. 1711--1717 (2016)

\bibitem{DDH:conf/ijcai/LinLT17}
Lin, J., Li, Z., Tang, J.: Discriminative deep hashing for scalable face image
  retrieval. In: IJCAI. pp. 2266--2272 (2017)

\bibitem{adversarial_fgir}
Lin, K., Yang, F., Wang, Q., Piramuthu, R.: Adversarial learning for
  fine-grained image search. In: ICME. pp. 490--495 (2019)

\bibitem{bcnn}
Lin, T.Y., RoyChowdhury, A., Maji, S.: Bilinear cnn models for fine-grained
  visual recognition. In: CVPR. pp. 1449--1457 (2015)

\bibitem{DH:conf/cvpr/LiongLWMZ15}
Liong, V.E., Lu, J., Wang, G., Moulin, P., Zhou, J.: Deep hashing for compact
  binary codes learning. In: CVPR. pp. 2475--2483 (2015)

\bibitem{DSH:conf/cvpr/Liu0SC16}
Liu, H., Wang, R., Shan, S., Chen, X.: Deep supervised hashing for fast image
  retrieval. In: CVPR. pp. 2064--2072 (2016)

\bibitem{Food101:conf/eccv/BossardGG14}
Lukas, B., Matthieu, G., Van~Gool, L.: Food-101 - mining discriminative
  components with random forests. In: ECCV. pp. 446--461 (2014)

\bibitem{AirCraft:journals/corr/MajiRKBV13}
Maji, S., Rahtu, E., Kannala, J., Blaschko, M.B., Vedaldi, A.: Fine-grained
  visual classification of aircraft. CoRR  \textbf{abs/1306.5151} (2013)

\bibitem{powerofasym}
Neyshabur, B., Srebro, N., Salakhutdinov, R.R., Makarychev, Y., Yadollahpour,
  P.: The power of asymmetry in binary hashing. In: NeurIPS. pp. 2823--2831
  (2013)

\bibitem{part_based_fgir}
Pang, C., Li, H., Cherian, A., Yao, H.: Part-based fine-grained bird image
  retrieval respecting species correlation. In: ICIP. pp. 2896--2900 (2017)

\bibitem{tripletloss}
Schroff, F., Kalenichenko, D., Philbin, J.: Face{N}et: A unified embedding for
  face recognition and clustering. In: CVPR. pp. 815--823 (2015)

\bibitem{SDH:conf/cvpr/ShenSLS15}
Shen, F., Shen, C., Liu, W., Shen, H.T.: Supervised discrete hashing. In: CVPR.
  pp. 37--45 (2015)

\bibitem{CUB:journals/CalTech/WBSWPB2011}
Wah, C., Branson, S., Welinder, P., Perona, P., Belongie, S.: The caltech-ucsd
  birds-200-2011 dataset  (2011)

\bibitem{eccv:hashing3}
Wang, G., Hu, Q., Cheng, J., Hou, Z.: Semi-supervised generative adversarial
  hashing for image retrieval. In: ECCV. pp. 469--485 (2018)

\bibitem{scda}
Wei, X.S., Luo, J.H., Wu, J., Zhou, Z.H.: Selective convolutional descriptor
  aggregation for fine-grained image retrieval. IEEE TIP  \textbf{26}(6),
  2868--2881 (2017)

\bibitem{piecewise}
Wei, X.S., Wang, P., Liu, L., Shen, C., Wu, J.: Piecewise classifier mappings:
  Learning fine-grained learners for novel categories with few examples. IEEE
  TIP  \textbf{28}(12),  6116--6125 (2019)

\bibitem{maskcnn}
Wei, X.S., Xie, C.W., Wu, J., Shen, C.: {Mask-CNN}: Localizing parts and
  selecting descriptors for fine-grained bird species categorization. Pattern
  Recognition  \textbf{76},  704--714 (2018)

\bibitem{SH:conf/nips/WeissTF08}
Weiss, Y., Torralba, A., Fergus, R.: Spectral hashing. In: NeurIPS. pp.
  1753--1760 (2008)

\bibitem{CNNH:conf/aaai/XiaPLLY14}
Xia, R., Pan, Y., Lai, H., Liu, C., Yan, S.: Supervised hashing for image
  retrieval via image representation learning. In: AAAI. pp. 2156--2162 (2014)

\bibitem{fgis_tmm}
Xie, L., Wang, J., Zhang, B., Tian, Q.: Fine-grained image search. IEEE
  Transactions on Multimedia  \textbf{17}(5),  636--647 (2015)

\bibitem{navigate}
Yang, Z., Luo, T., Wang, D., Hu, Z., Gao, J., Wang, L.: Learning to navigate
  for fine-grained classification. In: ECCV. pp. 420--435 (2018)

\bibitem{eccv:hashing1}
Yuan, X., Ren, L., Lu, J., Zhou, J.: Relaxation-free deep hashing via policy
  gradient. In: ECCV. pp. 134--150 (2018)

\bibitem{eccv:hashing5}
Yuan, X., Ren, L., Lu, J., Zhou, J.: Relaxation-free deep hashing via policy
  gradient. In: ECCV. pp. 134--150 (2018)

\bibitem{eccv:hashing4}
Zhang, J., Shen, F., Liu, L., Zhu, F., Yu, M., Shao, L., Heng~Tao, S.,
  Van~Gool, L.: Generative domain-migration hashing for sketch-to-image
  retrieval. In: ECCV. pp. 297--314 (2018)

\bibitem{macnn}
Zheng, H., Fu, J., Mei, T., Luo, J.: Learning multi-attention convolutional
  neural network for fine-grained image recognition. In: CVPR. pp. 5209--5217
  (2017)

\bibitem{crl}
Zheng, X., Ji, R., Sun, X., Wu, Y., Huang, F., Yang, Y.: Centralized ranking
  loss with weakly supervised localization for fine-grained object retrieval.
  In: IJCAI. pp. 1226--1233 (2018)

\bibitem{towardscrl}
Zheng, X., Ji, R., Sun, X., Zhang, B., Wu, Y., Huang, F.: Towards optimal fine
  grained retrieval via decorrelated centralized loss with normalize-scale
  layer. In: AAAI. vol.~33, pp. 9291--9298 (2019)

\bibitem{PDH:journals/tip/ZhuKZFT17}
Zhu, F., Kong, X., Zheng, L., Fu, H., Tian, Q.: Part-based deep hashing for
  large-scale person re-identification. {IEEE} TIP  \textbf{26}(10),
  4806--4817 (2017)

\end{thebibliography}
\end{document}